\documentclass[compsoc,conference,a4paper,10pt,times]{IEEEtran}
\IEEEoverridecommandlockouts

\usepackage{cite}
\usepackage{amsmath,amssymb,amsfonts}

\usepackage{amsthm}
\usepackage{thmtools}
\declaretheoremstyle[headfont=\bf]{normalhead}
\declaretheorem[style=normalhead]{definition}
\declaretheoremstyle[headfont=\bf]{normalhead}
\declaretheorem[style=normalhead]{example}

\usepackage{algorithmic}
\usepackage{graphicx}
\usepackage{textcomp}
\usepackage{bmpsize}
\usepackage[dvipsnames,svgnames]{xcolor}
\usepackage{lipsum}
\usepackage{hyperref}
\hypersetup{
    colorlinks=true,
    citecolor=black,
    linkcolor=black,
    urlcolor=black
}
\usepackage[linesnumbered, ruled,vlined]{algorithm2e}
\usepackage{stmaryrd}
\usepackage{calc}
\newenvironment{itemize*}%
  {\begin{itemize}%
    \setlength{\itemsep}{3pt}%
    \setlength{\parskip}{0pt}}%
  {\end{itemize}}
\usepackage{amsfonts}
\usepackage{amssymb}
\usepackage{textcomp}
\usepackage{colortbl}
\usepackage{xspace}
\usepackage{alltt}
\usepackage{color}
\usepackage{subfig}
\usepackage{listings}
\usepackage{url}
\usepackage{tikz-cd}
\usepackage{tikz}
\usetikzlibrary{arrows,shapes,shadows}
\usepackage{verbatim} 
\usepackage{mathtools}
\usepackage{graphicx}
\usepackage[T1]{fontenc}
\usepackage{enumitem}
\usepackage{fancybox}

\newcommand{\defeq}{\ensuremath{\mathbin{{:}{=}}}}

\newcommand{\mypara}[1]{\vspace{5pt}\noindent\textbf{#1.}}

\newcommand{\Figref}[1]{Figure\,\ref{#1}}

\newcommand{\la}{\ensuremath{\mathbin{\mathtt{{:}-}}}}

\newcommand{\solved}[1]{\ensuremath{#1^\lambda}}
\newcommand{\gsol}{\ensuremath{\solved{G}}\xspace}
\newcommand{\pI}{\ensuremath{\mathrm{I}}\xspace}
\newcommand{\pII}{\ensuremath{\mathrm{II}}\xspace}

\newcommand{\Flr}{\ensuremath{\mathsf{F}}}
\newcommand{\FlrStar}{\ensuremath{\Flr^*}}

\newcommand{\funM}{\ensuremath{\mathsf{M}}}

\newcommand{\true}{\textsc{true}\xspace}
\newcommand{\false}{\textsc{false}\xspace}
\newcommand{\undefined}{\textsc{undef}\xspace}

\newcommand{\ygreen}[1]{\textcolor{DarkGreen}{#1}}
\newcommand{\yred}[1]{\textcolor{DarkRed}{#1}}
\newcommand{\yyellow}[1]{\textcolor{DarkYellow}{#1}}

\newcommand{\xgreen} {{\ensuremath{\textcolor{DarkGreen}{\mathsf{green}}}}}
\newcommand{\xred}     {{\ensuremath{\textcolor{DarkRed}{\mathsf{red}}}}}
\newcommand{\xyellow}{{\ensuremath{\textcolor{DarkYellow}{\mathsf{yellow}}}}}
\newcommand{\xblue} {{\ensuremath{\textcolor{RoyalBlue}{\mathsf{blue}}}}}
\newcommand{\xorange} {{\ensuremath{\textcolor{DarkOrange}{\mathsf{orange}}}}}

\newcommand{\won} {{\ensuremath{\textcolor{DarkGreen}{\mathtt{W}}}}\xspace}
\newcommand{\lost}     {{\ensuremath{\textcolor{DarkRed}{\mathtt{L}}}}\xspace}
\newcommand{\drawn}{{\ensuremath{\textcolor{DarkYellow}{\mathtt{D}}}}\xspace}

\definecolor{DarkGreen}{rgb}{0,0.45,0}
\definecolor{DarkRed}{rgb}{0.8,0,0}
\definecolor{DarkYellow}{rgb}{0.6,0.6,0}
\definecolor{DarkGray}{rgb}{0.2,0.2,0.2}

\newcommand{\na}{\textcolor{gray}{\emph{n/a}}}
\renewcommand{\footnoterule}{%
  \kern -1pt
  \hrule width 0.4\columnwidth
  \kern 1pt
}

\newcommand{\plen}{\ensuremath{\mathsf{len}}\xspace}
\newcommand{\pval}{\ensuremath{\mathsf{val}}\xspace}
\newcommand{\mlen}{\ensuremath{\mathsf{len}}\xspace}
\newcommand{\mval}{\ensuremath{\mathsf{val}}\xspace}
\newcommand{\wonpr}{{\ensuremath{\textcolor{DarkGreen}{\mathtt{W_{\mathsf{pr}}}}}}}
\newcommand{\wonsc}{{\ensuremath{\textcolor{DarkGreen}{\mathtt{W_{\mathsf{sc}}}}}}}

\newcommand{\DTM}{\ensuremath{\mathsf{DTM}}}

\def\BibTeX{{\rm B\kern-.05em{\sc i\kern-.025em b}\kern-.08em
    T\kern-.1667em\lower.7ex\hbox{E}\kern-.125emX}}

\usepackage{listings} % Include the package at the beginning of your document

\begin{document}

%\title{On the Provenance of Games and Argumentation Frameworks}
\title{On the Structure of Game Provenance and its Applications}
\author{
\IEEEauthorblockN{Shawn Bowers} \smallskip
\IEEEauthorblockA{Department of Computer Science \\
\textit{Gonzaga University}\\
Spokane, USA \\
bowers@gonzaga.edu}
\and
\IEEEauthorblockN{Yilin Xia} \smallskip
\IEEEauthorblockA{School of Information Sciences \\
University of Illinois\\
Urbana-Champaign, USA \\
yilinx2@illinois.edu}
\and
\IEEEauthorblockN{Bertram Lud{\"a}scher} \smallskip
\IEEEauthorblockA{School of Information Sciences \\
University of Illinois\\
Urbana-Champaign, USA \\
ludaesch@illinois.edu}
}

\maketitle
%need remove afterwards
% \thispagestyle{plain}
% \pagestyle{plain}

%\newcommand{\pos}[1]{\ensuremath{\mathtt{#1}}}
%\newcommand{\pos}[1]{\ensuremath{\mathrm{#1}}}
\newcommand{\pos}[1]{\ensuremath{\mathsf{#1}}}

\newcommand{\winmove}{\pos{win}-\pos{move}}
\newcommand{\AF}{\ensuremath{\mathsf{AF}}}
\newcommand{\wm}{\ensuremath{\mathsf{WM}}}

\newcommand{\FIXME}{\marginpar{\yred{\textsc{fixme}}}}

\begin{abstract}
  Provenance in databases has been thoroughly studied for positive and for recursive queries, then
  for first-order (FO) queries, i.e., having negation but no recursion.  Query evaluation can be
  understood as a two-player game where the opponents argue whether or not a tuple is in the
  query answer.  This game-theoretic approach yields a natural provenance model for FO queries,
  unifying \emph{how} and \emph{why-not} provenance. Here, we study the fine-grain structure of
  {game provenance}.  A game $G=(V,E)$ consists of \emph{positions} $V$ and \emph{moves} $E$ and can be \emph{solved} by computing the well-founded model of a single,
  unstratifiable rule:
  \begin{math}
    \pos{win}(X) \la \pos{move}(X, Y), \neg \, \pos{win}(Y).
  \end{math}
  In the solved game \gsol, the {value} of a position $x\,{\in}\,V$ is either \textsc{won}, \textsc{lost}, or \textsc{drawn}. 
  This value is explained by the \emph{provenance} $\mathcal{P}(x)$, i.e.,
  certain (annotated) edges reachable from $x$. 
   We identify seven  \emph{edge types} that give rise to  new
  kinds of provenance, i.e., \emph{potential}, \emph{actual}, and \emph{primary}, and demonstrate  that ``\emph{not all moves are created equal}''. 
    We describe the new
 provenance types, % and their significance, 
 show how they can be computed while solving games, and discuss  applications, e.g., 
  for abstract argumentation frameworks.
  % We discuss applications of our refined
  % provenance model: (i) An \emph{argumentation framework} (AF) can be
  % understood as a game, so game provenance notions carry over to AF
  % graphs, increasing their explainability; (ii) query evaluation can
  % be understood as game, so queries expressed as games inherit the
  % various new provenance notions presented here.
\end{abstract}


\section{Introduction}\label{sec-1}
\noindent Evaluating a query $Q$ on a database instance $D$ yields an answer $A\,{=}\,Q(D)$. The
\emph{provenance} of a tuple $t\,{\in}\,A$ reveals additional information about the output, e.g., on
what input tuples $D_t\subseteq D$ the output $t$ depends (\emph{lineage}, \emph{why provenance}
\cite{cui2000tracing,buneman01:_why_where}), how $t$ was derived from those tuples $D_t$ (\emph{how
  provenance} \cite{green2007provenance,grigoris-tj-simgodrec-2012}) or, in case $t\,{\notin}\,A$,
why $t$ is missing from $A$ (\emph{why-not provenance}
\cite{chapman2009not,meliou2010complexity,herschel2017survey}).
The algebraic \emph{provenance semiring} framework \cite{green2007provenance} for positive
relational and for recursive queries brought order to the  earlier approaches, and won a
test-of-time award~\cite{green2017semiring}. The approach has  been extended to non-recursive
queries with negation
\cite{geerts2010database,kohler_first-order_2013,gradel_provenance_2020,glavic_data_2021}, but
approaches that combine recursion with negation remain underexplored. % largely unexplored.

% One application of this framework involves non-stratified rules applied to a directed graph $G=(V,E)$ with vertices $V$ and directed edges $E$ can be represented as $Q$.

%\newpage 
\mypara{Recursion through Negation} 
We study the provenance of \emph{unstratified} rules, in particular, the following query:
 \begin{equation}
  \pos{Q}(X) \la \pos{E}(X, Y), \neg \, \pos{Q}(Y). 
  \tag{$Q$}
\end{equation}
Note that  $Q$ is defined recursively \emph{through} negation and thus cannot be stratified. Given a graph $G=(V,E)$,  $Q$ allows different interpretations, each offering unique provenance insights: e.g.,  we can view $G$ as  a two-player game, where an edge $\pos E(x,y)$ means that a player may {move} a game pebble from  the current {position} $x$ to a  new position $y$, at which point it is the opponent's turn.
% The position $y$ is also called a \emph{follower} of the position $x$ (for $x,y\in V$).

\mypara{Games} To emphasize this game-theoretic view, we  can rename input and output relations and write $Q$ as follows:
% and $E$ as \emph{moves}, it aligns with combinatorial game theory, where the value of a position (won or lost) is determined by the winning strategy by an opponent:
\begin{equation}
  \pos{win}(X) \la \pos{move}(X, Y), \neg \, \pos{win}(Y). \tag{$Q_\pos{WM}$}
\end{equation}
\noindent Written in $Q_\pos{WM}$ (``win-move'') form, the rule has a  natural interpretation: Position 
% $x$ is (objectively!) won, if there is a move to a position $y$ which is \emph{not} won, i.e., lost (as it's now the opponent's turn). 
$x$ is \emph{winning} (short: a \emph{win}) if there is a
{move} to a position $y$ which is lost
for the opponent.\footnote{In draw-free games the complement of winning is losing.}
 % otherwise the complement of winning is losing or drawing.}  
If there are no  moves left to play, $x$ is lost.
The \emph{well-founded model} (WFM) \cite{van1991well}
of $Q_\pos{WM}$  captures the game semantics: $\pos{win}(x)$ is \true, \false, and \undefined in the WFM iff  $x$ is \textsc{won}, \textsc{lost}, and \textsc{drawn} in $G$, respectively.

\mypara{Argumentation Frameworks}
$Q$ also implements a meta-interpreter for solving \emph{argumentation frameworks}~\cite{dung1995acceptability}:  $V$ and $E$  now represent abstract \emph{arguments} and \emph{attacks}, respectively. An argument $x$ is \emph{defeated} if there is an attacking argument $y$ that is \emph{not} defeated (i.e.,  \emph{accepted}):\footnote{
% Similar to $Q_\pos{WM}$, after renaming relations---and here also: reversing edges---$Q_\pos{AF}$ is equivalent to $Q$, i.e., $\pos{E}(x,y) \Leftrightarrow   \pos{attacks}(y,x)$.
Like $Q_\pos{WM}$, rule $Q_\pos{AF}$ is equivalent to $Q$ after renaming relations and---in this
case---also reversing edges: $\pos{E}(x,y) \Leftrightarrow   \pos{attacks}(y,x)$.
}
 \begin{equation}
  \pos{defeated}(X) \la \pos{attacks}(Y, X), \neg \, \pos{defeated}(Y). 
  \tag{$Q_\pos{AF}$}
\end{equation}
The  WFM of $Q_\pos{AF}$ yields the \emph{grounded extension}  of a given argumentation framework (AF);  the stable models~\cite{gelfond_stable_1988} of  $Q_\pos{AF}$ yield all \emph{stable extensions} of AF \cite{dung1995acceptability,baroni_handbook_2018}. 

\mypara{Motivation} 
The reasons for studying  the (fine-grained) provenance structure of $Q$ are as follows:

\mypara{(1) Explanatory Power} Roughly speaking, \emph{solved games explain themselves}: Viewing $Q$
as a game $Q_\pos{WM}$ allows us to employ concepts from combinatorial game theory. The \emph{value}
of a position, e.g., is justified by concepts such as \emph{winning strategies} and the \emph{length}
of a position. These give rise to new provenance notions via different \emph{edge types}. In other
words, game-theoretic concepts provide additional ``provenance mileage'' and yield further insights
into why and how (or why-not and how-not) a result is derived.

\mypara{(2) Queries as Games} Query evaluation can be viewed as a game: e.g., all $n$-ary
\textsc{Fixpoint} queries can be brought into a normal form with an $n$-ary $Q$
\cite{kubierschky1995remisfreie,FKL97}. A similar construction, when applied to FO queries, yields a
unified approach for \emph{how} and \emph{why-not} provenance \cite{kohler_first-order_2013,lee_pug_VLDB2019},
extending the semiring approach \cite{green2007provenance}. Thus, although we focus on $Q$ (and its
``twins'' $Q_\pos{WM}$, $Q_\pos{AF}$), the concepts apply to other queries by generating
corresponding game normal forms.

\mypara{(3) Argumentation as a Game} The provenance structure derived from $Q_\pos{WM}$ applies,
\emph{mutatis mutandis}, also to $Q_\pos{AF}$, yielding an elegant and powerful explanatory
framework for the \emph{grounded extension} \cite{dung1995acceptability} of an argumentation
framework AF. Our approach explains why an argument is \emph{accepted}, \emph{defeated}, or
\emph{undecided}, much like our game provenance explains the values of positions in
games~\cite{ludascher_games_2023}.

\mypara{Outline and Contributions} %
The paper is organized as follows. Section~\ref{sec-2} presents some background on games, their
solutions, and our running example. In Section~\ref{sec-3}, we describe an iterative \emph {backward
  induction} algorithm for solving games. The \emph{alternating fixpoint procedure} (AFP),
originally described in \cite{van1993alternating}, operates similarly on $Q_\pos{WM}$. We then
introduce novel provenance notions for the positions in a game---\emph{potential}, \emph{actual},
and \emph{primary} provenance---introducing new labeling approaches and corresponding edge-type
annotations. Primary provenance extends our prior work \cite{kohler_first-order_2013} by providing a
more specific (minimal) explanation for the value of a position. We show how to compute the
different forms of provenance, including primary provenance and a complete set of new provenance
edge types and node labels for solved games. Section~\ref{sec-4} briefly discusses
applications for abstract argumentation frameworks and for query provenance under game normal-form
translations. We provide conclusions and plans for future work in Section~\ref{sec-5}.

\section{Background: Games on Directed Graphs}\label{sec-2}

\begin{figure}[t!]
  \centering
  \subfloat[Game Graph]{
    \includegraphics[width=0.48\columnwidth]{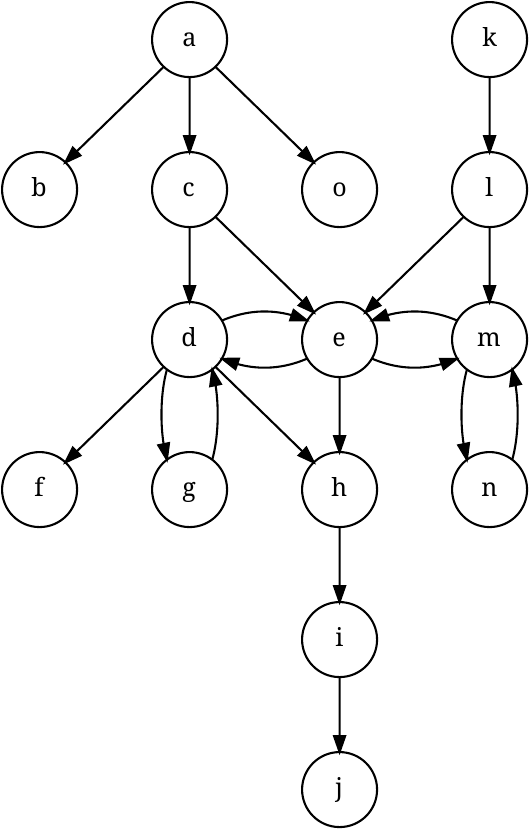}
    \label{fig:wm_unsolved}}
  ~~
  \subfloat[Solved Game]{
    \includegraphics[width=0.48\columnwidth]{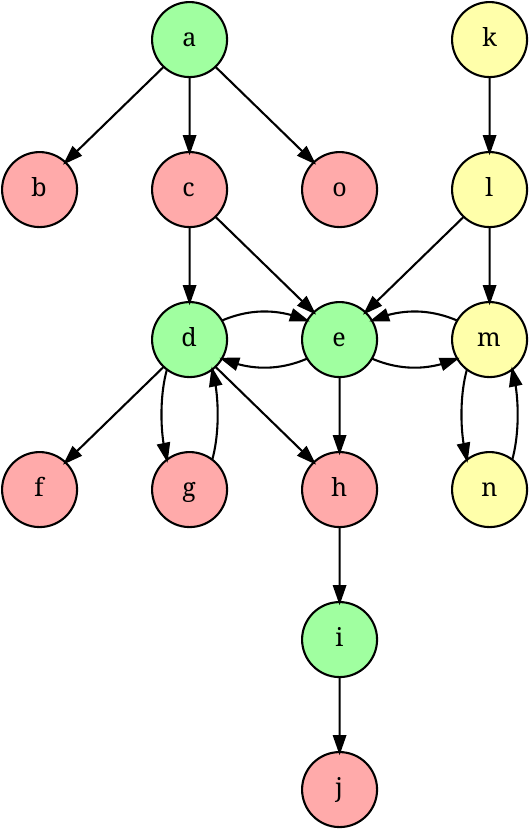}
    \label{fig:wm_nodes_only}}
  \caption{
    Game and its solution. The \emph{value} (label/color) of a position is either \won (\textsc{won}), \lost (\textsc{lost}), or \drawn (\textsc{drawn}).
  }\label{fig-running}
\end{figure}

% Outline:
% -- Positions and moves
% -- Take concepts / definitions from [KLZ13] (revise as needed)
% -- Introduce running example: (a) Move Graph (b) Well-founded Model
%    (\textcolor{red}{color coded for NODES only}!!)
% -- Theorem / Proposition from [FKL97] paper (WFS solves Games:
%    T/F/U for win(x) = win/lost/drawn of x)

\noindent By a \emph{game} we mean a finite, directed graph $G = (V,E)$. The vertices $V$ are  the \emph{positions}, the  edges $E \subseteq V \times V$ are the \emph{moves} of $G$. After choosing a start position $x_0$, Player~I and Player~II take turns, moving the game pebble along the edges $E$. 
A player who cannot move loses.\footnote{For example, in chess: \emph{Checkmate!}} 
% A player that forces their opponent into a
% losing position wins the game.  
Figure~\ref{fig-running} shows the  game we use as our running example.

\begin{definition}
  A \emph{play}
  $\pi_{x_0} = x_0 \stackrel{\pI}{\to} x_1 \stackrel{\pII}{\to} x_2 \stackrel{\pI}{\to} \cdots$ of a
  game $G = (V,E)$ is a sequence of moves $(x_i, x_{i+1})\,{\in}\,E$, starting at $x_0$ with
  Player\,I, and then players taking turns. % The two players
% of a win-move game take turns choosing moves within a play staring
% with Player~I, followed by Player~II, and so on.
  The \emph{length} $|\pi|$ of a play is given by its number of moves.
  $\pi$ is {\em complete} if it is of infinite length (repeating moves) or if it ends after $|\pi|$
  moves in a terminal (lost) position. % of the graph.
% Win-move games use so-called {\em normal play} rules where the player
% who cannot move {\em loses} the play $\pi$, while the previous player
% who made the last move {\em wins} $\pi$. 
\end{definition}

% NOTE: The following is troublesome since there is a "play" of
% infinite length in the example (d -> g)*
% ...
% A play $\pi$ of infinite length is a {\em draw}, which in finite
% game graphs implies that $E$ must have a cycle.

% \begin{example}
% Consider the play $\pi^1_{\pos e} = \pos e \to \pos d \to \pos f$ in
% Figure~\ref{fig:wm_unsolved} where \pos e is the start position for
% Player~I. In this play, Player~I moves to \pos d and Player~II moves
% to \pos f. Since I cannot move from \pos f, II
% wins. In the play
% $\pi^2_{\pos e} = \pos e \to \pos h \to \pos i \to \pos j$, however, Player~I
% moves to \pos h, Player~II moves to \pos i, and Player~I moves to \pos
% j. Since II cannot move, the play $\pi^2_{\pos e}$ is won for
% Player~I. This means that from position \pos e, the ``best'' move is
% $\pos e \to \pos h$, whereas the other moves available from {\pos e}
% are considered ``bad'': $\pos e \to \pos d$ results in a loss, while
% $\pos e \to \pos m$  results in a draw (if II sticks to
% $\pos m \to \pos n$). 
% \end{example}

\begin{example}
Let's consider some plays starting at $x_0\,{=}\,\pos{d}$ in \Figref{fig-running}: The best move for Player\,I from \pos d is to \pos f, leaving  Player\,II in a terminal node. Another winning play for I  is $\pos d \stackrel{\pI}{\to} \pos h \stackrel{\pII}{\to} \pos i  \stackrel{\pI}{\to} \pos j$. However, the move $\pos d \stackrel{\pI}{\to}\pos e$ is a \emph{blunder} as it leaves II in a winning position \pos e. The win can be  forced by II via $\pos e \stackrel{\pII}{\to} \pos h \stackrel{\pI}{\to} \pos i  \stackrel{\pII}{\to} \pos j$: Game over for I!
\end{example}

To determine the \emph{value} (\textsc{won}, \textsc{lost}, or
\textsc{drawn}) of a position, we assume  optimal play by both players. Thus, ``\emph{blunders}''
are ignored, such as moving from a winning position to a position that isn't lost for the opponent
(i.e., a position that is won or drawn {for the opponent}).

\begin{definition}[\textbf{Value of a Position}] A position $x$ is \textsc{won} if a player can force a win from $x$, independent of the opponent's moves; $x$ is \textsc{lost} if the opponent  can force a win; and $x$ is \textsc{drawn} if neither player can force a win.  
\end{definition} 

If a position $x$ is won, this means that the player moving from $x$ has a \emph{winning strategy}; if $x$ is lost, the opponent has a winning strategy; otherwise both players can delay a loss forever and force a draw by repetition.

% Restructure: 
%  Def 1 (Value of a Position). 
%  Def 2 (Optimal Play). No blunders, fastest win, slowest lost
%  Def 3 (Position Length). Remoteness
%  Def 4 (Solved Game: Node Labels) G^\star

\begin{definition}[\textbf{Solution of a Game}] The \emph{solution} of a game  is indicated by labeling (or coloring) each position with its true value: \won (\textsc{won}), \lost (\textsc{lost}), or \drawn (\textsc{drawn}).
\end{definition}

% The {\em solution} of a game graph $G = (V,E)$ consists of labeling
% each position with one of the values \won (\textsc{won}), \lost (\textsc{lost}), or \drawn
% (\textsc{drawn}). 
% The label of each position is given by a labeling (i.e.,
% solved position value) function
% $\pval : V \to \{\won,\lost,\drawn\}$.
% Specifically, a position $x$ is assigned the label \won if the player starting from $x$
% has a {\em winning strategy}: they can force a win from $x$ no matter
% how the opponent moves. Conversely, position $x$ is labeled \lost (lost) if the opponent can
% force a win no matter how the player starting at $x$ moves. In other words, {\em all} moves from
% $x$ result in a winning position for the opponent. If neither player
% can force a win, $x$ is labeled \drawn (a draw). 

\begin{example}
Figure~\ref{fig:wm_nodes_only}
shows the labeled solution for the game in Figure~\ref{fig:wm_unsolved}. Node colors indicate labels.
% (\xgreen\ $\sim$ \won, \xred\ $\sim$ \lost, and \xyellow\ $\sim$ \drawn).
 \end{example}

 \mypara{Solving Games with $Q_\pos{WM}$}
A game can be solved by evaluating $Q_\pos{WM}$ under the \emph{well-founded
   semantics}~\cite{van1991well}.  If this is implemented via the \emph{alternating fixpoint
   procedure} (AFP)~\cite{van1993alternating}, one obtains an increasing sequence of underestimates
 $U_1 \subseteq U_3 \subseteq U_5 \dots$
%$V_1^\won\subseteq V_3^\won \subseteq V_5^\won \dots$
converging to 
 the set of \true atoms $U^\omega$
% $V^\won$
from below, and a decreasing sequence of
overestimates 
$O_0 \supseteq O_2 \supseteq O_4 \dots$ 
%$V_0^\lost \supseteq V_2^\lost \supseteq V_4^\lost \dots$ 
converging to $O^\omega$, the \true or \undefined (undefined) atoms from
above. Thus, the atoms in the ``gap'' $O^\omega\setminus U^\omega$ have the third truth-value \undefined, while atoms not in  $O^\omega$ are \false.

On $Q_\pos{WM}$, AFP performs a \emph{backward induction} known from game theory. It also yields  additional game-theoretic information, notably the \emph{length} of a position (or \emph{remoteness function} \cite{smith_graphs_1966}).
% (i.e., the fastest way that a player can force a win, or the longest delay that a player can achieve before losing).
This  in turn provides additional insight into the provenance of the value of a position.

% \begin{definition}[\textbf{Optimal Play}] Assume Player~I has a winning strategy from a position $x$. Under \emph{optimal play}, Player~I will try to win from $x$ as quickly (i.e, in as few moves) as possible, while Player~II will try to delay defeat as long (i.e., in at most moves) as possible.   
% \end{definition}

% \begin{definition}[\textbf{Length of a Position} \cite{smith_graphs_1966}] The \emph{length} of a position $x$ is: the minimum number of moves to a win if $x$ is \textsc{won}; the maximum number of moves to a defeat if $x$ is \textsc{lost}; and $\infty$ if $x$ is \textsc{drawn}.  
% %Note that terminal nodes are lost by definition and have a length of 0.
% \end{definition}

\begin{definition}[\textbf{Length of a Position}] 
For a position $x$, $\plen(x)$ denotes the \emph{length} of $x$, i.e., the number of moves  necessary to force a win from $x$ (if $x$ is \textsc{won}), and the number of moves one can delay losing (if $x$ is \textsc{lost}). If $x$ is \textsc{drawn}, its length is $\infty$. This assumes that  players  try to win as quickly or to lose as slowly as possible. 
\end{definition}

\noindent Summarizing, we view solved games as labeled graphs:

\begin{definition}[\textbf{Solved Game with Length}] The \emph{solution} of game $G$ \emph{with
    lengths} is a labeled graph $G^\lambda = (V, E, \lambda)$ where $\lambda$ consists of two
  labeling functions $\pval_\lambda$ and $\plen_\lambda$:
\begin{itemize*}
\item $\pval_\lambda : V \to \{\won,\lost,\drawn\}$, returning position values; and
\item $\plen_\lambda : V \to \mathbb{N} \cup \{\infty\}$, returning position lengths.
\end{itemize*}
\end{definition}

\noindent We often omit $\lambda$ from the labeling functions $\pval$ and $\plen$.

\newcommand{\Prov}{\ensuremath{\mathcal P}}
\newcommand{\Pprov}{\ensuremath{\Prov_{\mathsf{pt}}}}
\newcommand{\Aprov}{\ensuremath{\Prov_{\mathsf{ac}}}}
\newcommand{\Mprov}{\ensuremath{\Prov_{\mathsf{pr}}}}
\newcommand{\Sprov}{\ensuremath{\Prov_{\mathsf{sc}}}}

\section{The Structure of Game Provenance}
\label{sec-3}

\noindent We  introduce and study different kinds of provenance for games and show how, using a variant of a simple backward induction algorithm, the \emph{primary provenance} of a game can be computed ``on the fly'' while solving a game.

% In this section, we begin by introducing the three types of provenance
% including potential provenance, actual provenance and minimum
% provenance. The basic algorithm used to solve \wm~graph, with a
% detailed example and visualization to illustrate how it works. By
% introducing the strategies for edge coloring, we further delve into
% the advanced algorithm which provide more provenance mileage to
% understand the solved \wm~graph.

\subsection{Solving Games Iteratively}
\label{sec-3-walkthrough}

\newcommand{\GR}{\textit{GR}\xspace}
\newcommand{\RR}{\textit{RR}\xspace}

\noindent  A  game can be solved by iterating  two labeling rules:

% \begin{itemize*}
% \item  $x$ is \textsc{lost} (\lost) if
% $\forall (x,y)\in E$:   $y$ is \textsc{won} \hfill  (\RR)
% \item $x$ is \textsc{won} (\won) 
% if $\exists\, (x,y)\in E$ s.t.\   $y$ is \textsc{lost} \hfill (\GR) 
% \end{itemize*}

\newcommand{\becomes}{\ensuremath{\mathbin{:=}}}

\begin{itemize*}
\item  $\pval(x) \becomes \lost$ if $\forall\, (x,y)\in E$:   $\pval(y)=\won$  \hfill  (\RR)
\item $\pval(x) \becomes \won$ if $\exists\, (x,y)\in E$ s.t.\   $\pval(y)=\lost$ \hfill (\GR) 
\end{itemize*}

% \begin{itemize}
% \item  $x$ is \textsc{lost} (\lost) if
% $\forall (x,y)\in E$:   $y$ is \textsc{won} \hfill  (\RR)
% \item $x$ is \textsc{won} (\won) 
% if $\exists\, (x,y)\in E$ s.t.\   $y$ is \textsc{lost} \hfill (\GR) 
% \end{itemize}

% The AFP-based approach for solving games effectively applies these two rules (one after the other) in a ``bottom up'' fashion by navigating moves in their reverse direction starting with the terminal positions of a game graph.
% Specifically, the algorithm starts by looking for
% terminal positions, which are then labeled as lost (\lost) since they have
% no outgoing moves (the red rule). The positions with at least one
% outgoing move to a lost position are then labeled as won (\won)
% according to the green rule. The positions with all of their outgoing
% moves landing on winning nodes are then labeled as lost (\lost)
% according to the red rule. This process repeats, switching between the
% green and red rules, until no additional positions can be labeled
% (i.e., a fixpoint is reached). Any remaining unlabeled positions of
% the game graph are then labeled as a draw (\drawn). We refer to each application of one of the two rules in the backward induction as a \emph{step}.  

\mypara{Red Rule}  (\RR) states that a position $x$ is  \textsc{lost} (\lost) if \emph{all} followers $y$ of $x$  have already been  labeled \textsc{won} (\won): No matter to which follower $y$ of $x$ a player moves to, the opponent can force a win from $y$. 

\mypara{Green Rule} (\GR) %, on the other hand,  
states that a position $x$ is  \textsc{won} (\won) if there \emph{exists} a follower $y$ of $x$ that has already been labeled \textsc{lost} (\lost): A player can thus choose to move from $x$ to  such a $y$, leaving the opponent in a lost position.

Initially, only (\RR) is applicable, and only to nodes without followers, i.e., terminal nodes (the
$\forall$-condition is vacuously satisfied). As soon as lost nodes are found, (\GR) becomes
applicable, yielding new won nodes, after which (\RR) may be triggered etc.\ until a \emph{fixpoint}
is reached.  Any remaining unlabeled positions of the graph are then labeled \textsc{drawn}
(\drawn). We refer to each complete application of one of the rules in this backward induction as a
\emph{step}.

\begin{figure*}[ht!]
  \centering
  \subfloat[Step 0 (Red Rule)]{
    \includegraphics[width=.5\columnwidth]{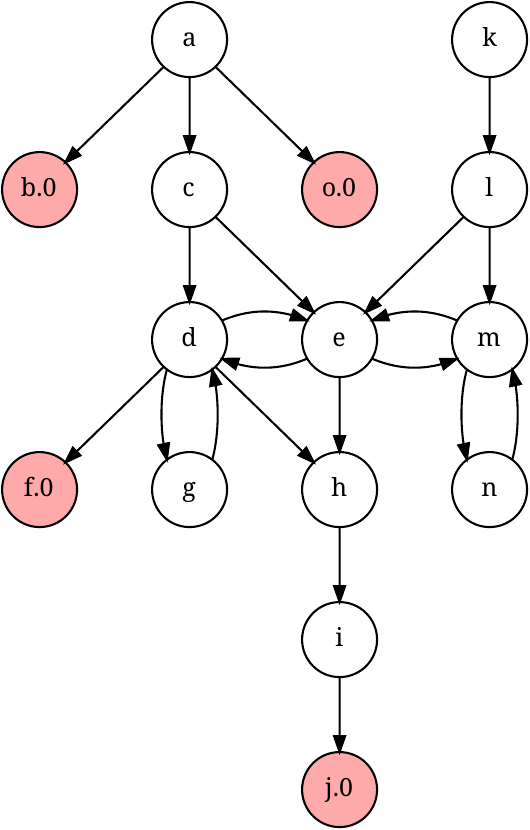}
    \label{fig:wm_state_1}}
  \hfill%\hfill\hfill
  \subfloat[Step 1 (Green Rule)]{
    \includegraphics[width=.5\columnwidth]{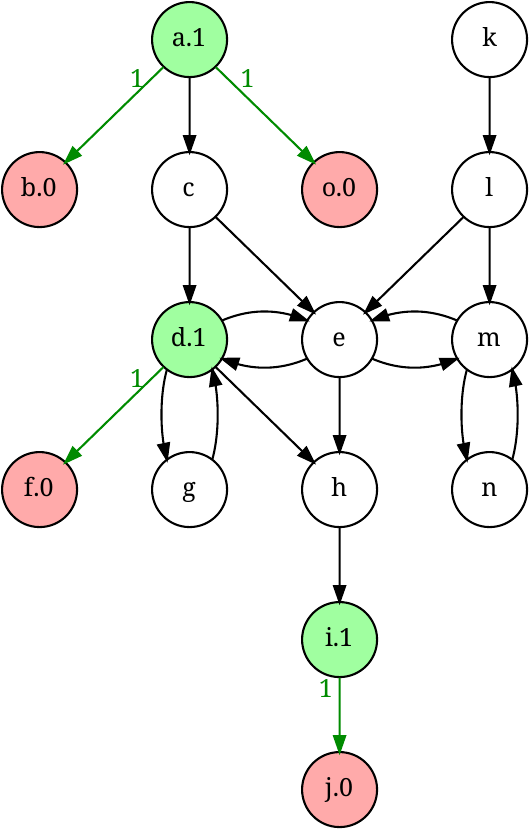}
    \label{fig:wm_state_2}}
  \hfill%\hfill\hfill
  \subfloat[Step 2 (Red Rule)]{
    \includegraphics[width=.5\columnwidth]{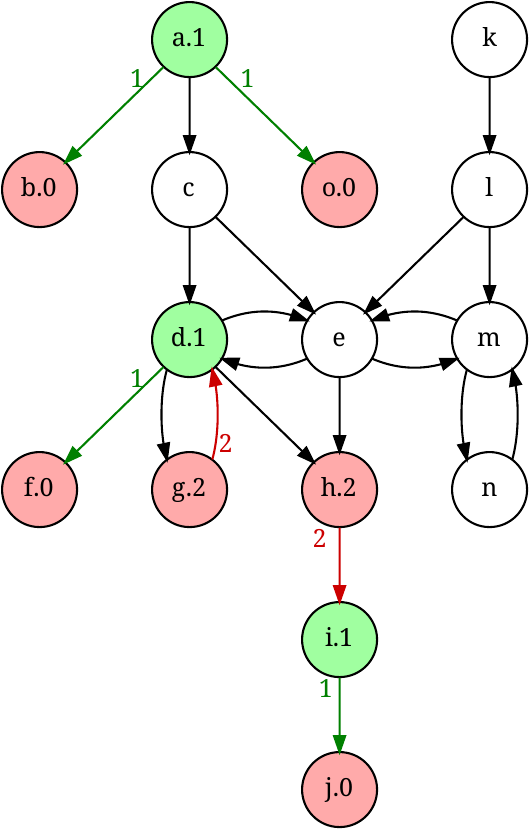}
    \label{fig:wm_state_3}} \\%[6pt]
  \subfloat[Step 3 (Green Rule)]{
    \includegraphics[width=.5\columnwidth]{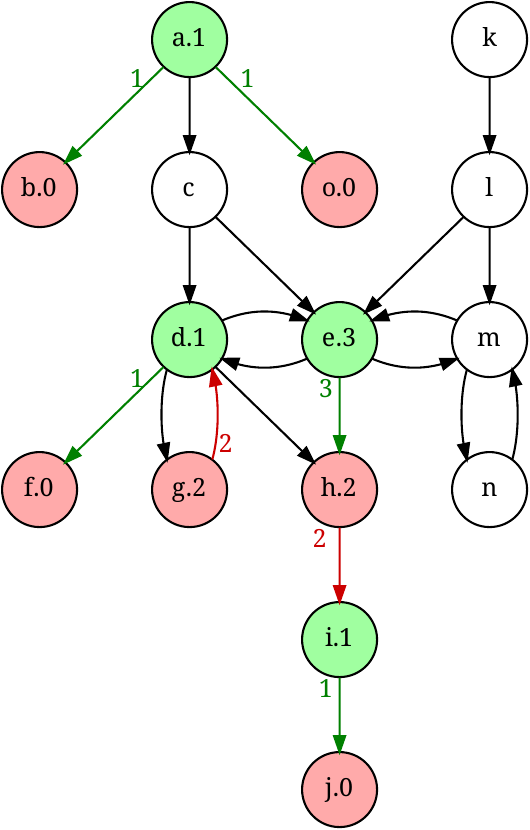}
    \label{fig:wm_state_4}}
  \hfill%\hfill\hfill
  \subfloat[Step 4 (Red Rule)]{
    \includegraphics[width=.5\columnwidth]{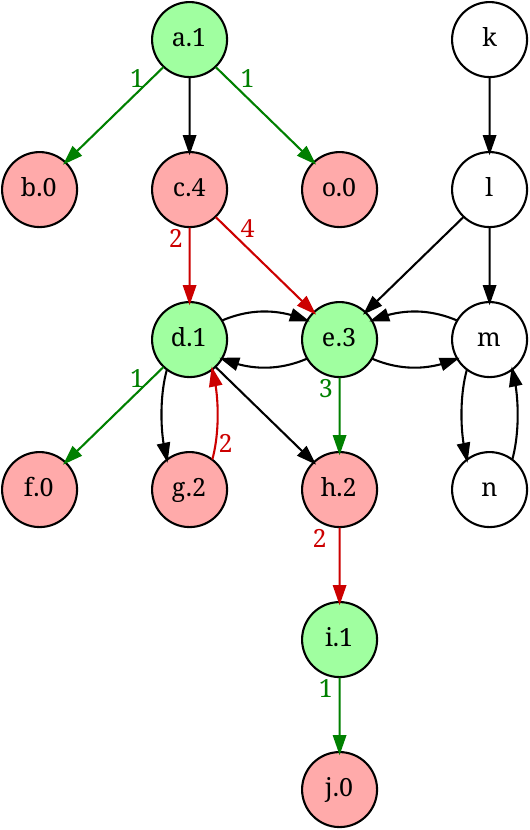}
    \label{fig:wm_state_5}}
  \hfill%\hfill\hfill
  \subfloat[Step 5+ (Full Provenance)]{
    \includegraphics[width=.5\columnwidth]{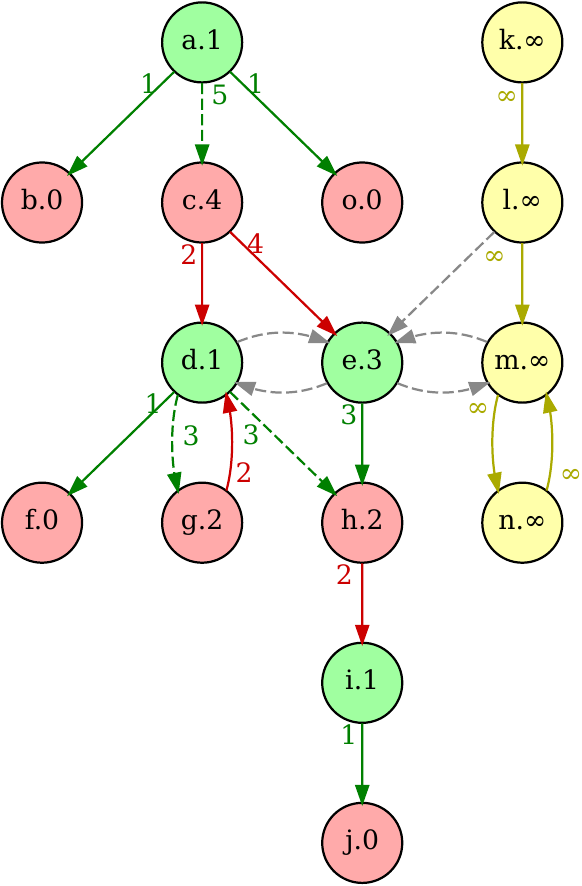}
    \label{fig:wm_complete}} 

  \caption{(a) The  graph in the upper left shows the initial
    state of the computation: Terminal nodes $x$ are
    immediately \textsc{lost} and colored  \xred; (b) Immediate predecessors $x$ of lost nodes  are  \textsc{won}, shown in \xgreen; (c), (d), (e) show  subsequent states according to the application of the Red Rule (\RR) and the Green Rule (\GR), respectively. Once the
    fixpoint is reached, all remaining nodes are known to be \textsc{drawn} and colored 
    \xyellow. % Move types (colors, lengths) explained in the text. % The color of edges can be
    % deduced based on the color of nodes it linked.
    % color-labeling as in (e).
  }
  \label{fig:animation}
\end{figure*}

\begin{example}
 Figure~\ref{fig:animation} traces this iterative algorithm
using the graph from Figure~\ref{fig-running}.
Some additional node and edge labels depicted in Figure~\ref{fig:animation}  are described further below.
\begin{itemize}
\item {\em Step 0 (\RR)}: Figure~\ref{fig:wm_state_1} shows the result of the initial step of
  labeling terminal nodes \lost.  Positions \pos b, \pos f, \pos j, and \pos o are each immediately
  lost (and colored \xred) since they have no outgoing moves. Each of these positions are also
  labeled with the step number 0.
\item {\em Step 1 (\GR)}: Figure~\ref{fig:wm_state_2} shows the
  result of applying (\GR) for the first time. Positions \pos a,
  \pos d, and \pos i are labeled \won (and colored \xgreen) since they
  each have at least one move to a lost node. The new \won nodes are labeled with step number 1.
\item {\em Step 2 (\RR)}: Figure~\ref{fig:wm_state_3} depicts the
  result of the second firing of  (\RR). Both positions \pos g and
  \pos h are  labeled \lost:   \pos g
  is lost since it only has one outgoing move to  \pos d, which has label \won. Similarly,  \pos h is lost
  since it only has one  move, i.e.,  to the winning position
  \pos i. Both \pos g and \pos h are labeled with step number 2.
\item {\em Step 3 (\GR)}: In Figure~\ref{fig:wm_state_4} we see the
  result of the next firing of (\GR):  \pos e is labeled \won and gets step number~3.
\item {\em Step 4 (\RR)}: Applying (\RR) again results in \pos c being labeled \lost (Figure~\ref{fig:wm_state_5}), receiving step number 4: Both followers of \pos c are known to be won (\won).  
\item {\em Step 5 (\GR)}: Applying (\GR) again does not change the labeling, so a fixpoint is reached. 
\end{itemize}
After reaching a fixpoint, the remaining 
positions are labeled \drawn (\textsc{drawn}, colored \xyellow) as shown in
Figure~\ref{fig:wm_complete}. The  step number of a position provides a record of when the position's value first became known; it corresponds to the \emph{length} of a position defined above.  \textsc{drawn} nodes receive the step number $\infty$. 
Additional edge labels and colors  shown in Figure~\ref{fig:wm_complete} are explained further below.
\end{example}

\begin{figure*}[ht!]
  \centering
  \subfloat[Potential provenance of \pos d]{
\includegraphics[width=.55\columnwidth]{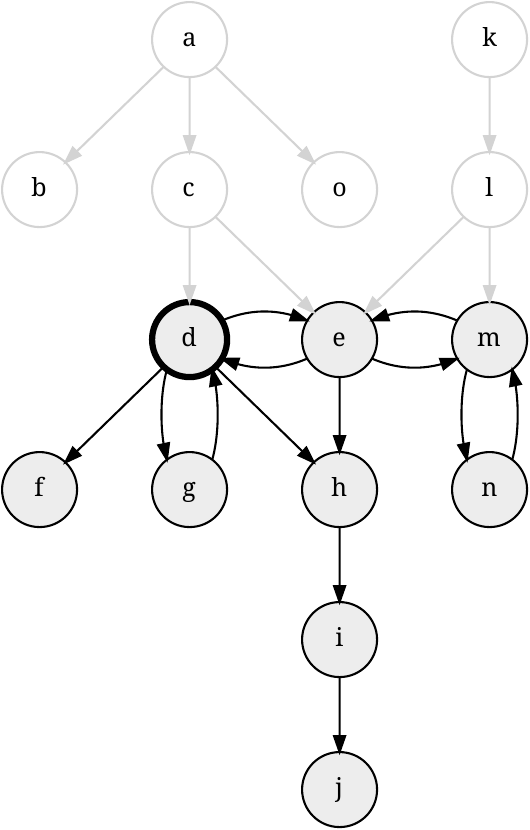}
    \label{fig:pprov}}
    \hspace{1cm}
  \subfloat[Actual provenance of \pos d]{
    \includegraphics[width=.55\columnwidth]{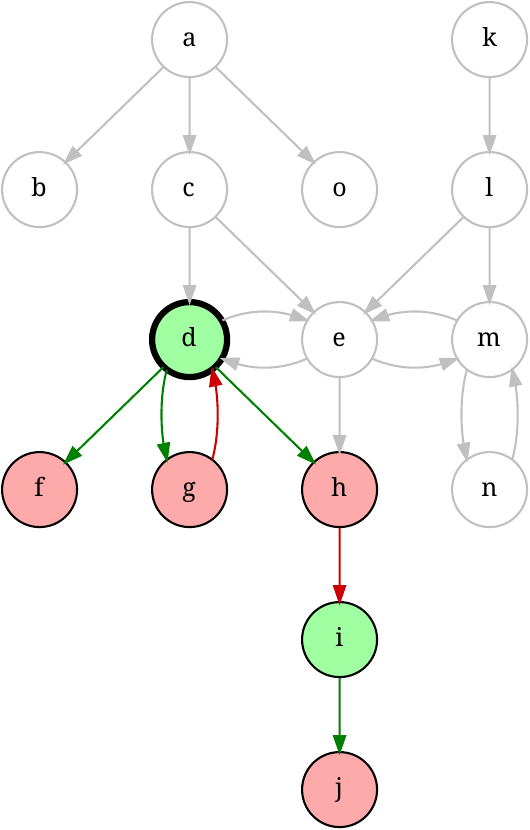}
    \label{fig:aprov}}
    \hspace{1cm}
  \subfloat[Primary provenance of \pos d]{
    \includegraphics[width=.55\columnwidth]{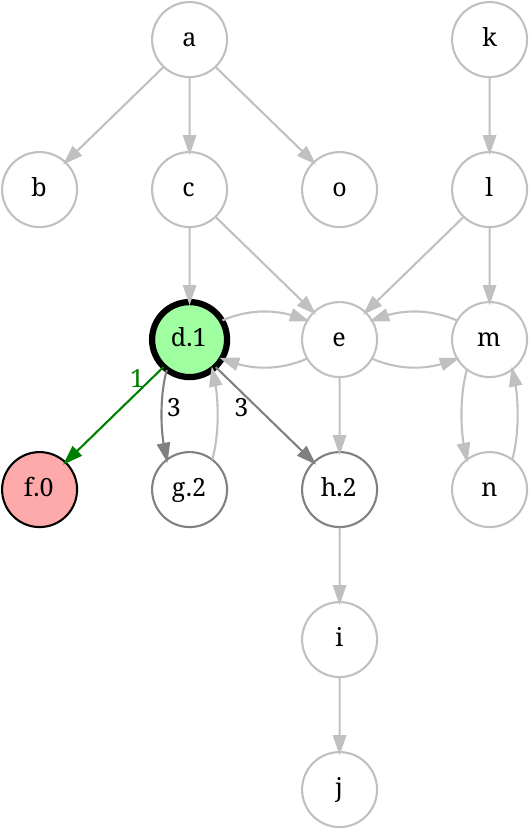}
    \label{fig:mprov_d}}
  % \subfloat[The primary provenance of \pos c]{
  %   \includegraphics[width=.5\columnwidth]{figures/prov_min_c.pdf}
  %   \label{fig:mprov_c}} 
  \caption{(a) The \emph{potential provenance} $\Pprov(\pos d)$ are all moves reachable from \pos d; (b) The \emph{actual provenance} $\Aprov(\pos d)$ avoids \emph{blunders} (``bad moves'') like $\pos d\to\pos e$; and (c) The \emph{primary provenance}
    $\Mprov(\pos d)$ uses \emph{minimal-length} winning moves only.}
  \label{fig:three_provs}
\end{figure*}

%\subsection{Degrees of Game Provenance}
\subsection{Potential, Actual, and Primary Provenance}

\noindent We consider the {\em provenance} \Prov\ of the value of a position $x\in V$ in a game $G$
to be an explanation of why $x$ has a particular value. The explanation of the value of a position
$x$ is closely tied to the complete plays $\pi_x$ that can be started from $x$ and is represented by
a labeled subgraph rooted at $x$.  Below we define three novel types of
provenance---\emph{potential}, \emph{actual}, and \emph{primary provenance}---each providing more
specific (i.e., often smaller) subgraphs explaining the value of $x$.

%\mypara{Potential Provenance} 
As described in Section~\ref{sec-2}, the value of a position $x$ is based on the possible plays
$\pi_x$ \emph{starting} at $x$.\footnote{Unlike plays $\pi$ \emph{into} $x$ or \emph{ending} in $x$:
  They do \emph{not}  impact $x$'s value.}  % In particular, a position
% $x$ is won if a player who is currently moving can force a win no
% matter how its opponent moves, $x$ is lost if the opponent of the
% player who is currently moving can force a win, and if neither player
% can force a win $x$ is drawn. 
Thus, the value of $x$ can \emph{only} depend on the moves reachable from $x$ in the game.  We refer to this notion of provenance as potential provenance.

\begin{definition}[\textbf{Potential Provenance}] The \emph{potential provenance} $\Pprov(x)$ of $x$ is the subgraph reachable from $x$.
\end{definition}

% We denote the potential provenance of a position $x$ in a game $G$ as $\Pprov(x)$ $(= \Pprov^G(x))$. 

\begin{definition}[\textbf{Followers}]
  The \emph{followers} of a position $x$ are its immediate successors:
  $\Flr(x) \becomes \{y \mid (x,y) \in E\}$. 
For sets $S\subseteq V$, define $\Flr(S) \becomes \bigcup_{x\in S}\Flr(x)$. 
% With this, define
%   $\FlrStar(x) =  \{x\}\cup\Flr(x)\cup\Flr(\Flr(x))\cup\Flr(\Flr(\Flr(x)))\cdots$
Now define the \emph{closure} as
  $\FlrStar(x) \becomes  \{x\}\cup\Flr(x)\cup\Flr^2(x)\cup\Flr^3(x)\cdots$
\end{definition}

\noindent 
The potential provenance $\Pprov(x)$ is thus the subgraph of $G$ with nodes $\FlrStar(x)$
and edges $\{(u,v) \in E \mid u, v \in \FlrStar(x)\}$.

\begin{example}
Figure~\ref{fig:pprov} gives the potential provenance
$\Pprov(\pos d)$ of position \pos d for the game of
Figure~\ref{fig:wm_unsolved}. As shown, $\Pprov(\pos d)$ defines a
subgraph (via the gray nodes) of $G$ (with corresponding move
vertices) containing all possible, but not necessarily optimal, plays starting at \pos d. The
potential provenance does not rely on $G$ being solved (labeled), which is
emphasized in Figure~\ref{fig:pprov} by coloring the corresponding
positions gray.    
\end{example}

% \mypara{Actual Provenance} 
While potential provenance \Pprov\ captures
the typical notion of provenance as all dependencies of a node, it can
overestimate the justification for why a position has a particular
value in a game. For example, a position $x$ is won if it has at least
one move that forces the opponent to lose, regardless of the moves the
opponent makes (GR). However, $x$ may still have outgoing moves that if
followed would be blunders for the player (e.g., by allowing the opponent
to win or to draw). Similarly, a position that is a draw for a player
may have an outgoing move that relinquishes the draw by allowing their
opponent to win. While such moves can be contained in a position's potential provenance, they do not determine the value of the position. We refer to this notion of considering only moves that can determine the value of a position as \emph{actual provenance}.  The actual provenance of a position is obtained directly from a solved (labeled) game extended with additional move labels. 

\begin{definition}[\textbf{Provenance Edges}] We extend to edges the labeling functions $\lambda$ for solved games $\gsol(V, E, \lambda)$:
% , the labeling functions $\lambda = \{ \pval, \plen \}$ are extended to moves such that:  
\begin{itemize*}
\item $\mval_\lambda : E \to \{\won, \lost, \drawn\}$ is a partial function: 
% giving the value of each move such that:
\begin{displaymath}
  \mval_\lambda(x,y) \becomes
  \left\{
    \begin{array}{ll}
      \won & \mbox{if } \pval_\lambda (x) = \won \mbox{ and } \pval_\lambda (y) = \lost \\
      \lost & \mbox{if } \pval_\lambda (x) = \lost \mbox{ and } \pval_\lambda (y) = \won \\
      \drawn & \mbox{if } \pval_\lambda (x) = \drawn \mbox{ and } \pval_\lambda (y) = \drawn \\
    \end{array}
  \right .
\end{displaymath}
\item $\mlen_\lambda  : E \to \mathbb{N} \cup \{\infty\}$ is a partial function:
\begin{displaymath}
  \mlen_\lambda(x,y) \becomes
  \left\{
    \begin{array}{ll}
      1 + \plen_\lambda(y) & \mbox{if } \mval_\lambda(x,y) \in \{\won,\lost\} \\
      \infty & \mbox{if } \mval_\lambda(x,y) = \drawn \\
    \end{array}
  \right .
\end{displaymath}
\end{itemize*}
\end{definition}

% \footnote{Losing positions cannot have ``bad'' moves since all
%   moves out of the position, by definition, are to winning positions.}

\noindent  We define actual provenance based on the labels (colors) of move edges in the solved game.

\begin{definition}[\textbf{Actual Provenance}] $\Aprov(x)$, the \emph{actual provenance} of a position $x$, is the subgraph reachable from $x$ by only following \won-, \lost-, and \drawn-labeled edges. 
\end{definition}

\begin{example}  
Figure~\ref{fig:aprov} shows the actual provenance $\Aprov(\pos d)$ of
position \pos d in the running example. Note that the actual provenance is a proper subgraph of the potential provenance. In particular,
none of the plays through \pos e are considered, since the move $\pos d \to \pos e$ is a blunder. In Figure~\ref{fig:aprov}, the relevant positions and moves are
colored, representing the fact that the edge-labeling function \mval was used in obtaining \pos d's actual provenance.
\end{example}

\emph{Optimal play} means selecting  a move that guarantees the {fastest} (i.e., minimal-length) win among the possible \emph{winning} moves of a won position. Conversely, the best strategy for a player in a losing position is to select a maximal-length \emph{delaying} move that loses the slowest. 

Optimal play is assumed by the backward induction described in
Section~\ref{sec-3-walkthrough}. Note that the value of a winning
position $x$ becomes first known  after one of its successor positions $y$ is discovered to be lost. For instance, position \pos d in Figure~\ref{fig:aprov} ultimately has three winning moves. 
However, the winning move discovered first, $\pos d \to \pos f$, is also the fastest win. This move is
 the one used by the Green Rule (\GR) to determine the winning
value of \pos d. Thus,  $\pos d \to \pos f$ is a
\emph{primary provenance} edge. It is also the
 edge used by the AFP-based evaluation of $Q_\pos{WM}$ to determine that \pos d is winning. Equivalently, it is part of the optimal strategy when playing from
the winning position \pos d. The other two winning moves $\pos d \to \pos g$ and $\pos d \to \pos h$ 
 are \emph{secondary provenance}. Primary and secondary edges are determined directly from the lengths of moves of the solved game.

\begin{definition}[\textbf{Primary Provenance}] $\Mprov(x)$, the \emph{primary provenance}  of a position $x$, is the subgraph reachable from $x$ via \lost, \drawn, and \emph{minimum-length} \won edges.
\end{definition}

\noindent Note the asymmetry in the  previous definition: The value of a winning position $x$  depends on the existence of a follower $y$ which is lost (for the opponent). However, the length of $x$ is determined solely by the subset of  lost followers that have \emph{minimal length} (optimal play means winning as fast a possible). 
% Thus, the primary provenance only includes the subset of such followers of minimal length. 
In contrast, the value of a lost position is dependent on \emph{all} followers being winning for the opponent, i.e., it is \emph{not} sufficient to include only the subset of maximal-length delaying moves. A drawn position depends on the existence of followers that are drawn themselves (a drawn position $x$ cannot have lost followers, otherwise $x$ wouldn't be drawn but won). 

% We denote the primary provenance of a position $x$ in a game $G$ as $\Mprov(x) (= \Mprov^G(x))$. 

% Assume $\gsol$ is the solution of $G$ and $E_{\pos{sc}} \subseteq E$ is the set of winning secondary (i.e., non-minimal) moves. Let $G_{\textsf{pr}} = (V, E_\pos{pr})$ such that $E_\pos{pr} = E_\pos{ac} \setminus E_\pos{sc}$ consists of all labeled edges of $E$ except the secondary edges. The primary provenance of $x$ can be defined as its potential
% provenance over $G_{\textsf{pr}}$: 
% $\Mprov^G(x) = \Pprov^{G_{\textsf{pr}}}(x)$. 
% The primary provenance is a subgraph of the actual provenance since $E_{\textsf{pr}} \subseteq E_{\textsf{ac}}$. Thus, $\Mprov^G(x) \subseteq \Aprov^G(x) \subseteq \Pprov^G(x)$.

\begin{figure}[t!]
  \centering
    \includegraphics[width=0.55\columnwidth]{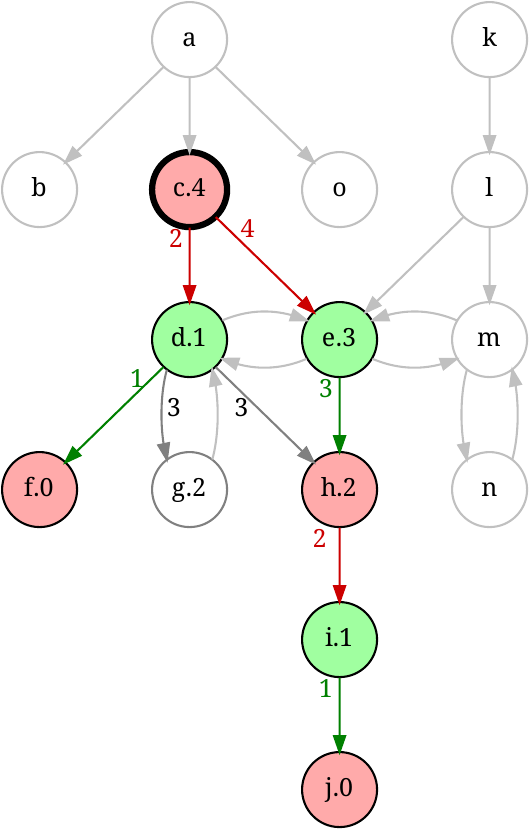}
  \caption{$\Mprov(\pos c)$ contains \emph{all} delaying moves (\pos d's value depends on all of them) but only fastest winning moves.}
  \label{fig:mprov_min_c}
\end{figure}

\begin{algorithm*}[ht!]
  \algsetup{linenosize=\small}
  \small
  \SetKwData{Left}{left}\SetKwData{This}{this}\SetKwData{Up}{up}
  \SetKwFunction{Union}{Union}\SetKwFunction{FindCompress}{FindCompress}
  \SetKwInOut{Input}{Input}\SetKwInOut{Output}{Output}
  \Input{Game graph $G = (V,E)$}
  \Output{Solved game with \emph{primary} provenance $G^{\lambda_{\mathsf{pr}}} = (V,E,\lambda_{\mathsf{pr}})$}
  \BlankLine
  % Initially we don't know any won positions
%  $V_\mathit{won} \defeq \emptyset$  {\footnotesize\tcp*[r]{Initially we don't know any won positions}}
  $V_{\mathit{won}} \defeq \emptyset$  {\footnotesize\tcp*[r]{Initially we don't know any won positions}}

  % All sinks are lost
%  $V_\mathit{lost} \defeq \{ x\in V \mid \Flr(x) = \emptyset \}$ {\footnotesize\tcp*[r]{but all sinks are lost}}
  $V_{\mathit{lost}} \defeq \{ x\in V \mid \Flr(x) = \emptyset \}$ {\footnotesize\tcp*[r]{but all sinks are lost}}
  % the state of all sinks are 0
  % $node\_color(x) \defeq \xred$ for all $x\in V_\mathit{lost}$  \tcp*[r]{ all lost positions are colored red.} 
  $\plen(x) \defeq 0$ for all $x\in V_\mathit{lost}$  {\footnotesize\tcp*[r]{their length is 0.}} 
  
  % iterate the Green and Red rules
  \Repeat{$V_\mathit{won}$ and $V_\mathit{lost}$ change no more} {    
    
    % for uncolored nodes only
    \For(\tcp*[f]{\footnotesize Apply the (\GR) rule to unlabeled nodes}){$x \in V \setminus (V_\mathit{won} \cup V_\mathit{lost})$} {
    %\For{$x \in V \setminus (V_\mathit{won} \cup V_\mathit{lost})$ {\footnotesize\tcp*[r]{Consider unlabeled nodes only.}}} {
      %  get the following nodes whose color are known
      $F_\mathit{lost} \defeq \Flr(x) \cap V_\mathit{lost}$ \;
      % some nodes followers are lost
      \If{$F_\mathit{lost} \neq \emptyset$ } %\tcp*[r]{\textbf{\emph{some}} $y\in\Flr(x)$ is  lost, so $x$ is \textbf{\emph{won}}}
      {
        $V_\mathit{won} \defeq V_\mathit{won} \cup \{x\}$ ; ~ $\pval(x) \defeq \won$ {\footnotesize\tcp*[r]{\textbf{\emph{some}} $y\in\Flr(x)$ are  lost, so $x$ is \textbf{\emph{won}} (\xgreen)}}
        $\plen(x) \defeq 1 + \min \{ \plen(y) \mid y \in F_\mathit{lost} \}$ {\footnotesize\tcp*[r]{\textbf{\emph{shortest}} win}}
        % $\forall y \in F_\mathit{lost} ~ \lambda(x,y) \defeq (\won, \length(x))$ ;
        \For{$y \in F_\mathit{lost}$
        } {
          $\mval(x,y) \defeq \wonpr ; ~
          \mlen(x, y) \defeq 1+\plen(y)$ 
        }
      }% if end      
    }      
    
    % for uncolored nodes only
    \For(\tcp*[f]{\footnotesize Apply the (\RR) rule to unlabeled nodes}){$x \in V \setminus (V_\mathit{won} \cup V_\mathit{lost})$} {
      %  get the following nodes whose color are known
      $F_\mathit{won} \defeq \Flr(x) \cap V_\mathit{won}$ \;
      % all followers are won
      \If{$\Flr(x) = F_\mathit{won}$}  
      {

        $V_\mathit{lost} \defeq V_\mathit{lost} \cup \{x\}$ ; ~ $\pval(x) \defeq \lost$ {\footnotesize \tcp*[r]{\textbf{\emph{all}} $y\in\Flr(x)$ are won, so $x$ is \textbf{\emph{lost}} (\xred)}}
      
        $\plen(x) \defeq 1 + \max \{ \plen(y) \mid y \in F_\mathit{won} \}$ {\footnotesize\tcp*[r]{\textbf{\emph{longest}} delay}}

        \For{$y \in F_\mathit{won}$} {
          $\mval(x,y) \defeq \lost ; ~ \mlen(x,y) \defeq 1+\plen(y)$
        }      
      }
    }
  } 
  \caption{Solve Game with Primary Provenance}
  \label{algo:improved_algorithm}
\end{algorithm*}

\begin{example}
Figure~\ref{fig:mprov_d} shows the primary provenance $\Mprov(\pos d)$ of position \pos d, which is a proper subgraph of its actual provenance. In particular, only the move $\pos d\,{\to}\,\pos f$ is considered from \pos d, since it has a length of 1, whereas the other two winning moves $\pos d\,{\to}\,\pos g$ and $\pos d\,{\to}\,\pos h$ both have length~3. 
In Figure~\ref{fig:mprov_min_c}  the primary provenance $\Mprov(\pos c)$ of the lost position \pos c is shown. In this case,  both of its delaying moves to \pos d and \pos e are included in the primary provenance. In contrast,  in Figure~\ref{fig:mprov_d}, the moves $\pos d\,{\to}\,\pos g$ and $\pos d\,{\to}\,\pos h$ were not included. Finally, the  move $\pos e\,{\to}\,\pos h$ is included in the primary provenance  as it is the only winning move from~\pos e.
\end{example}

% Given a solved game graph $G^\lambda(V,E)$ as above, let $E_\pos{sc}$
% be the secondary moves of the winning positions in $V$. Note that
% $E_\pos{sc}$ can be obtained directly for winning positions $x \in V$
% by finding all of the non-minimal \mlen moves $x \to y$ such that y is
% a lost position.

% \begin{figure*}[ht!]
%   \centering
%   \subfloat[The potential provenance of \pos d]{
%     \includegraphics[width=.55\columnwidth]{figures/prov_potential.pdf}
%     \label{fig:pprov}}
%   \hfill%\hfill\hfill
%   \subfloat[The actual provenance of \pos d]{
%     \includegraphics[width=.55\columnwidth]{figures/prov_actual.pdf}
%     \label{fig:aprov}}
%   \hfill%\hfill\hfill
%   \subfloat[The primary provenance of \pos d]{
%     \includegraphics[width=.55\columnwidth]{figures/prov_min.pdf}
%     \label{fig:mprov}} \\%[6pt]

%   \caption{
%     % color-labeling as in (e).
%   }
%   \label{fig:three_provs}
% \end{figure*}

\subsection{Computing Primary Provenance}
\label{sec_computing_primary_prov}

% \mypara{Solving Games with Provenance Extensions}
\noindent Algorithm~\ref{algo:improved_algorithm} extends the backward induction described in Section~\ref{sec-3-walkthrough} by adding steps for labeling positions and moves via \mval and \mlen functions. The result  is a labeled  graph $G^{\lambda_{\mathsf{pr}}} = (V, E, \lambda_\mathsf{pr})$ consisting of all \won and \lost \emph{position} values and lengths, and all \lost and all {primary} $\ygreen{\won_{\textsf{pr}}}$ \emph{edge} values and lengths. The core of the algorithm
repeatedly applies the Red Rule (\RR) followed by the Green Rule (\GR). The sets $V_{\mathit{won}}$ and $V_{\mathit{lost}}$
are used to track the won and lost positions, respectively:
% The algorithm works as follows.
\begin{itemize}
\item Initially, no won positions are known (line 1). All terminal positions are lost (line 2). The length is set to 0 and assigned to each terminal node (line 3).
\item The algorithm repeatedly applies the (\GR) and (\RR) rules (lines 5--18) until a fixpoint is reached, i.e., when no additional won or lost nodes are found.
\item At each \emph{round}, i.e., (\GR) followed by (\RR), only positions $x$ that have not yet been assigned a value (lines 5 and 12) are considered. $F_\mathit{lost}(x)$ and $F_\mathit{won}(x)$ are the followers of $x$ that are known to be lost and  won, respectively (lines 6 and 13).
\item Lines 7--11 implement (\GR): If $x$ has a lost follower (line 7), $x$ is assigned \won (line 8) and its length (line~9). 
The moves $x\to y$ for losing followers $y\in F_\mathit{lost}$ are assigned  \wonpr\ (line 11). Since the winning moves were just found, they have minimal length (assigned in line~11) and are part of the primary provenance.
\item Lines 14--18 implement (\RR): If \emph{all} followers of $x$ are winning  (line 14), $x$ is assigned \lost (line 15) and its length is set (line 16). Each  move  $x\to y$  is assigned value \lost and its length (line 18).
\end{itemize}

\noindent The algorithm adds 1 to the minimum (maximum) length of a lost (a won) follower, respectively. Note that the value and length of a move $x\to y$  is determined by the value and length of $x$'s follower $y$.
An alternative, but equivalent, approach would be to
maintain an explicit step number (as in 
Section~\ref{sec-3-walkthrough}) and use this number to assign lengths to positions and moves. The only exception is the computation of the length for losing (delaying) moves, which has to be computed as shown in line 16.

% For winning positions, the step
% number can also be used for their primary provenance \mlen
% labels. However, for losing positions, the specific \plen labels of
% their followers must be used to assign \mlen move labels as in line 16
% of the algorithm.

% The \mlen labels on moves provide immediate information regarding the
% ``win-fast, lose-slow'' optimal plays of a player from a particular
% game position. For instance, position \pos a in
% Figure~\ref{fig:wm_complete} has two ``win fast'' moves (hence the \mlen
% label 1): $\pos a \to \pos b$ and $\pos a \to \pos o$. Both of these
% moves represent the primary provenance of position \pos a (denoted by
% solid move edges), and as a result, are the moves that were
% specifically used to determine that position \pos a is a winning
% position. Conversely, \pos c has a single ``lose slow'' move
% $\pos c \to \pos e$ whose corresponding losing play $\pi_{\pos c}$
% involves four moves (as given by the \mlen label 4) that go through
% position \pos c and end at position \pos j.

\mypara{Goal-Oriented Provenance Computation} Algorithm~\ref{algo:improved_algorithm} computes the primary provenance $G^{\lambda_{\mathit{pr}}}$ of all positions in the game simultaneously. If the provenance of specific nodes is needed on large graphs, an initial pre-processing step can be used to reduce the size of the input graph. 
In particular, the potential provenance $\Pprov(x)$ of position $x$ can be computed first, i.e., the subgraph of $G$ rooted at $x$. This subgraph can then be used as input to the algorithm, resulting in the primary provenance $\Mprov(x)$. 
Further optimizations can be applied: e.g., \cite{Zinn2012WC} describe a \emph{distributed} and \emph{disorderly} evaluation of $Q_\mathsf{WM}$ that can solve  a large game graph without the need for a central compute node to have a copy of the complete graph. 

\mypara{Computing Full Provenance} 
The example in Figure~\ref{fig:animation} (a--e)  was used to illustrate the iterative solution of games in Section~\ref{sec-3-walkthrough}. Unlike that base method, Algorithm~\ref{algo:improved_algorithm} also computes values and lengths of moves $x\to y$. The final result of the algorithm is depicted in Figure~\ref{fig:wm_state_5}.
To  compute the \emph{full provenance} of our  example (Figure~\ref{fig:wm_complete}), additional post-processing steps are needed:
\begin{itemize}
\item for all unlabeled $x$: $\pval(x) \becomes \drawn$ and
$\plen(x) \becomes \infty$; 
\item for all unlabeled moves $x \to y$ of type $\won \to \lost$:  
$\mval(x,y) \becomes \wonsc$ (\emph{secondary}, i.e., slow-winning move)  

$\mlen(x,y) \becomes 1 +  \plen(y)$; 
\item for all unlabeled $x \to y$ of type $\drawn \to \drawn$:
$\mval(x,y) \becomes \drawn$ and $\mlen(x,y) \becomes \infty$.
\end{itemize}

% Figure~\ref{fig:wm_complete} depicts the final result. 

\mypara{Blunders $\neq$ Provenance} Figure~\ref{fig:wm_complete} contains three edge types not yet discussed: \emph{blunders}. These are ``bad moves'' that do not contribute to the provenance of a position $x$, since although they are part of the \emph{potential} provenance of $x$, they are not part of $x$'s \emph{actual} provenance. 

\begin{figure}[t!]
  \centering
\includegraphics[width=0.9\columnwidth]{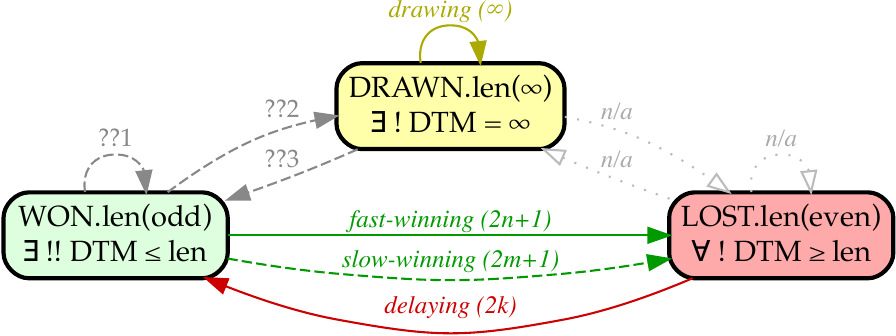}
    \caption{\textbf{Seven Edge Types.} 
The \emph{edge type} of $x\to y$ is determined by the {values} of $x$ and $y$: It can be (fast or slow) \ygreen{\emph{winning}}       (\ygreen{\textsc{won}}$\to$\yred{\textsc{lost}}), \yred{\emph{delaying}}
      (\yred{\textsc{lost}}$\to$\ygreen{\textsc{won}}),
      or
      \yyellow{\emph{drawing}}
(\yyellow{\textsc{drawn}}$\to$\yyellow{\textsc{drawn}}).
   %   Winning moves are further divided into {\em primary}  (solid) and {\em secondary} (dashed).  
   The three types of \emph{blunders} (??1, ??2,  ??3)  give away an advantage to the opponent and are not part of any provenance. Dotted ``ghost edges''  (\na) 
      cannot exist due to the semantics of games. Adopting chess terminology, the \emph{distance-to-mate} (\DTM) for a 
\ygreen{\textsc{won}} position $x$ is at most $\plen(x) = 2n+1$; for a 
\yred{\textsc{lost}} $y$, mate can be delayed at least $\plen(y) = 2k$ moves.}
    \label{fig:edge_types}
\end{figure}

For example, a move $x\to y$ of type $\won \to \won$ is a blunder of type-1, depicted as $\ygreen{\textsc{won}} \stackrel{\mathrm{??1}}{\longrightarrow} \ygreen{\textsc{won}}$ in Figure~\ref{fig:edge_types}: The player moving from $x$ had a winning position but blundered victory away and gave it to the opponent. Clearly, such moves do not contribute to the value (and length) of a position $x$. 
Similarly, moves of type-2 and type-3, i.e.,  
$\ygreen{\textsc{won}} \stackrel{\mathrm{??2}}{\longrightarrow} \yyellow{\textsc{drawn}}$ 
and
$\yyellow{\textsc{drawn}} \stackrel{\mathrm{??3}}{\longrightarrow} \ygreen{\textsc{won}}$ 
are blunders, yielding a draw where a win was possible, and a loss where a draw was possible, respectively.

Figure~\ref{fig:edge_types} shows all seven \emph{move types} that can occur in a solved game: \emph{fast} and \emph{slow winning} $\won \to \lost$ (primary and secondary, resp.); \emph{delaying} $\lost \to \won$ (losing); and \emph{blundering} $\won \to \won$ (type-1), $\won \to \drawn$ (type-2), and $\drawn \to \won$ (type-3).

\newcommand{\RNum}[1]{\lowercase\expandafter{\romannumeral #1\relax}}

\subsection{The Regular Structure of Game Provenance}

\noindent Looking at the solved game in Figure~\ref{fig:wm_nodes_only}, one notices that certain types of edges are absent, e.g., there are \emph{no moves} from a lost  node to another node that is lost or drawn. 
Similarly, there always \emph{is} a move from a won node to a lost node and from a drawn node to another drawn node. These are not accidents.

The \emph{type graph} in Figure~\ref{fig:edge_types} summarizes the provenance structure of solved games, i.e., of node values and edge types. The seven edge types can be split into provenance-relevant moves (\emph{winning}, \emph{delaying},  \emph{drawing}), with edge labels \won, \lost, and \drawn, respectively, and provenance-\emph{irrelevant} moves (blunders of type-1, type-2, and type-3). 

The \emph{winning} moves can be further subdivided into \emph{fast-winning}, which are part of the primary provenance, and \emph{slow-winning}, which constitute secondary provenance. 

Figure~\ref{fig:edge_types} also shows that there are three types of ``ghost moves’’ that cannot exist in a solved game graph: e.g., a position would not be lost if there were a move to a lost or to a drawn position. Similarly, a drawn $x$ can never have a move to a lost follower $y$, otherwise $x$ would be winning rather than being drawn.

\mypara{The Structure of Game Provenance} 
In the following, we sketch a simple pattern-based method that reveals the regular provenance structure of solved games. In addition to yielding insights into games in general (and into other formalisms that can be reduced to games),  and explaining and justifying the values of positions in particular, we can also use this pattern-based approach to explore and query provenance.

\mypara{Provenance Paths} 
Consider a game $G=(V,E)$. Its move relation $E$ induces a function $\funM^R$ (for \emph{move paths matching $R$)}. Here $R$ is a regular expression over an alphabet $\{\wonpr, \wonsc, \won, \lost, \drawn\}$ of edge labels. 

\begin{definition}
The expression $\funM^R(x)$ evaluates to the minimal subgraph $G'\subseteq G$ rooted at $x$ such that all paths
\begin{displaymath}    
    x\stackrel{\ell_1}{\rightarrow} x_1 \stackrel{\ell_2}{\rightarrow} x_2 \cdots \stackrel{\ell_n}{\rightarrow}x_n
\end{displaymath}
in $G$, whose concatenated labels $\ell_1 \ell_2 \cdots \ell_n$ match the regular expression $R$, are also matched by $G'$.
\end{definition}

 In this way, the parameter $R$ specifies a \emph{regular path query} (RPQ) \cite{wood12}, but unlike an RPQ which returns a set of nodes, $\funM^R(x)$ returns a subgraph definable by an RPQ.

\mypara{Actual Provenance via RPQs} 
The actual provenance $\Aprov(x)$ of a position $x$ can be defined using $\funM^R$:
\begin{displaymath}
\Aprov(x) \becomes
  \left\{
    \begin{array}{ll}
      \funM^{\won.(\lost.\won)^*}(x) & \mbox{if } \pval_\lambda(x) = \won \\
      \funM^{(\lost.\won)^*}(x) & \mbox{if } \pval_\lambda(x) = \lost\\
      \funM^{\drawn^+}(x) & \mbox{if } \pval_\lambda(x) = \drawn \\
    \end{array}
  \right .
\end{displaymath}
Here, \won is defined as  $(\wonpr | \wonsc)$.

\mypara{Primary Provenance via RPQs}
The primary provenance $\Pprov(x)$ of a position $x$ is defined by: 
\begin{displaymath}
\Mprov(x) \becomes
  \left\{
    \begin{array}{ll}
      \funM^{\wonpr.(\lost.\wonpr)^*}(x) & \mbox{if } \pval_\lambda(x) = \won \\
      \funM^{(\lost.\wonpr)^*}(x) & \mbox{if } \pval_\lambda(x) = \lost\\
      \funM^{\drawn^+}(x) & \mbox{if } \pval_\lambda(x) = \drawn \\
    \end{array}
  \right .
\end{displaymath}

\noindent  RPQs for actual provenance were also described in \cite{kohler_first-order_2013}. 

% \noindent Consider the actual provenance in Figure~\ref{fig:aprov} and the 
% full provenance depicted in Figure~\ref{fig:wm_complete}. Close inspection reveals that 

% Figure~\ref{fig:edge_types} summarizes the different types of 
% edge labelings used 

% in Figure~\ref{fig:wm_complete}, where ``bad'' edges
% are colored brown and are dashed, primary winning edges are colored
% green and are solid, and secondary edges are colored green and are
% dashed.

\section{Applications of Game Provenance}
\label{sec-4}
% \textbf{*** TODO: shall we add other applications, link back to the
% main conference? (not full sure what we can add tho...) Otherwise it
% probably should be called "application in AF and discussion"?}

% With the solved game and degrees of granularity regarding game
% provenance explained in Section~\ref{sec-3}, this section provides two
% application scenarios where we apply the refined provenance structure
% to explain Argumentation Framework and help Query Evaluation.

\noindent We consider two separate areas where game provenance can be  applied. We first discuss applications to abstract
argumentation frameworks and then briefly discuss the application to the provenance of queries. 

% \subsection{Argumentation Framework Interpretation}a

\mypara{Argumentation through the Lens of Games} There is a direct correspondence between game graphs $G = (V, E)$ and abstract argumentation frameworks (AFs) \cite{ludascher_games_2023}. 
In particular, positions in an AF graph represent abstract {\em arguments} and
edges represent \emph{attack} relationships: An edge $x \to y$ in
$E$ means that argument $x$ is {\em attacked-by} argument $y$. The typical convention in AFs is
to read the {\em attacked-by} relation in the opposite direction,
i.e.,  saying that argument $y$ {\em attacks} $x$.  

\begin{figure}[t!]
  \centering
\includegraphics[width=0.6\columnwidth]{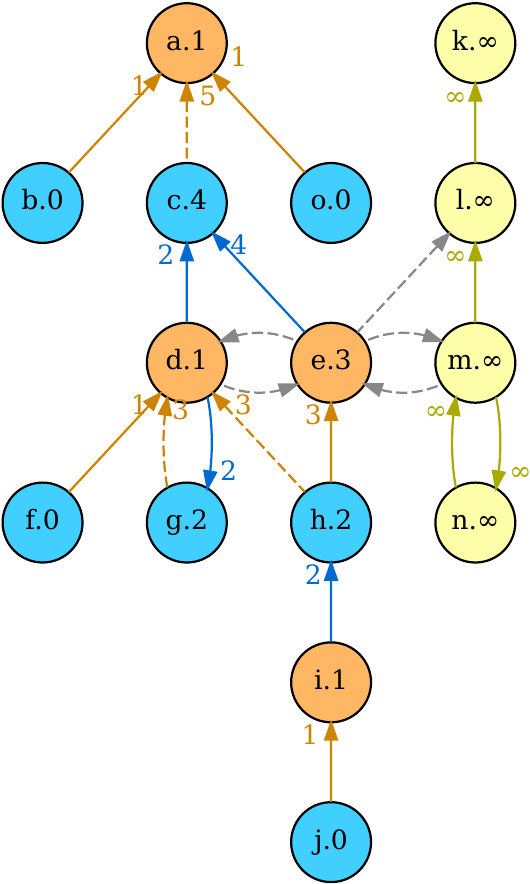}
  \caption{Solved argumentation framework AF (grounded extension), derived from the well-founded model of $Q_\pos{AF}$. Here, \emph{accepted} arguments are
    \xblue, \emph{defeated} arguments  \xorange, and \emph{undecided} arguments  \xyellow. The RPQ-definable provenance structure of solved games,  the seven edge types, etc.\ all carry over---\emph{mutatis-mutandis}---to AF.}
  \label{fig:atk_solved}
\end{figure}

Figure~\ref{fig:atk_solved}
shows the running example as a solved AF graph, where the
edges are shown in the \emph{attacks} direction. Nodes and edges are colored in a different style used for AF graphs. Given an argumentation framework, one
of the goals is to find sets of arguments that can be jointly 
  accepted. An argument is  \emph{accepted} if it has no
attackers,  or if  all of its attackers are attacked by at least one
accepted argument.
An argument is said to be {\em defeated} if it is attacked by at least
one accepted argument.\footnote{Somewhat counter to intuition, \emph{accepted} (\emph{defeated}) arguments correspond to  \emph{lost} (\emph{won}) positions in a game, respectively \cite{ludascher_games_2023}.}

A set of accepted arguments is called an {\em extension}. An important
class of extensions for  \emph{skeptical} (versus
\emph{credulous}) reasoning are the {\em grounded extensions} \cite{dung1995acceptability}, which
 correspond exactly to the well-founded model of the query $Q_\pos{AF}$. 
% :
%  \begin{equation}
%   \pos{defeated}(X) \la \pos{attacks}(Y, X), \neg \, \pos{defeated}(Y). 
%   \tag{$Q_\pos{AF}$}
% \end{equation}
That is, $\pos{defeated}(x)$ is \true, \false, and
\undefined in the WFS iff argument $x$ is \emph{defeated}
(\xorange), \emph{accepted} (\xblue), or neither, which is also referred to as \emph{undecided} (\xyellow) in
AFs. Thus, the algorithms in Section~\ref{sec-3} can be used
(after reversing edges and renaming relations) to find the grounded extension
of an AF. The labels generated can then be used to explain why a given
argument is accepted or defeated via the actual $\Aprov(x)$ and
primary $\Pprov(x)$ provenance of a node $x$.

\usetikzlibrary{arrows, automata}

\mypara{Queries through the Lens of Games} In
\cite{kohler_first-order_2013}, FO (first-order) queries are translated
into a {\em game normal form} for unifying \emph{how} and \emph{why-not} 
provenance. In particular, the game $G_{Q(D)}$ of a first-order query
$Q$ (expressed in non-recursive Datalog syntax) simulates an SLDNF evaluation of $Q$ over an input database
$D$. The positions of the query evaluation game $G_{Q(D)}$ correspond, e.g., to rule firings and ground literals. The game is constructed in such a way that the position that encodes a possible query answer is \emph{won} iff it is an answer to $Q(D)$, and \emph{lost} otherwise. Since FO queries are non-recursive, the resulting game graph is acyclic and thus contains no \emph{drawn} positions. 

The how and why-not provenance of a query
answer corresponds to our notion of actual provenance. Primary provenance refines the work in \cite{kohler_first-order_2013}, and can
be used to find more specific derivations of a query answer (or
non-answers in the case of why-not provenance). 
%according to SLDNF resolution. 
This in turn has applications, e.g., for reducing the
provenance of an answer (when multiple derivations of increasing
length are possible) as well as for query debugging. 

\section{Discussion and Future Work} \label{sec-5}

\noindent We have studied in detail the problem of explaining why a
position in a game is won, lost, or drawn, respectively, and revealed the fine-grained, regular structure of  provenance in solved games. This game provenance has immediate 
applications, e.g., for argumentation frameworks and for queries expressed in
a game normal form.  Our approach considers the provenance of a game
position $x$ as the (relevant) moves and positions that are reachable from a starting position $x$. 

By considering the structure of game provenance that
results from solving a game, we extend  prior work
\cite{kohler_first-order_2013,lee_pug_VLDB2019,ludascher_games_2023} and identify
three increasingly more specific levels of provenance: \emph{potential}, \emph{actual},
and \emph{primary} provenance of a node. We also show how the actual and  \emph{full provenance} labelings of a game can be computed, and how the standard
approach for solving games (via backward induction) can be extended
to capture primary provenance. The latter provides a 
specific ``fastest'' justification  for why a position is won or lost.
An open source software demonstration of our approach is available \cite{xia_game-provenance_tapp_2024}. In future
work, we plan to develop additional tools for analyzing and visualizing the provenance of games and argumentation frameworks. We also plan to
extend our approach for \emph{credulous} reasoning, i.e., employing stable models \cite{gelfond_stable_1988} and stable AF extensions \cite{dung1995acceptability}.

% Games have emerged as a crucial method in advancing our understanding
% of database queries. By revisiting and refining existing provenance
% models and algorithms, we've developed a more nuanced provenance
% model, introducing varying levels of granularity to game
% provenance. This approach aims to enhance our ability to explain the
% reasons behind won, lost, or drawn positions within a game. This
% refined structure also has implications for both the interpretation of
% argumentation frameworks and the query evaluation. 
% In the future, we plan to introduce such approach to stable models in
% game theory as well as preferred extensions in argumentation
% framework.

%%% Local Variables:
%%% mode: latex
%%% TeX-master: "game-provenance-tapp"
%%% End:

%\newpage

\bibliographystyle{IEEEtran}
\bibliography{TAPP24/Argumentation-and-Games}

\appendices
\end{document}